\pdfoutput=1

\documentclass[11pt]{article}

\usepackage{acl2023}

\usepackage{times}
\usepackage{latexsym}

\usepackage[T1]{fontenc}

\usepackage[utf8]{inputenc}

\usepackage{microtype}
\usepackage[nolist]{acronym} 
\usepackage{booktabs}
\usepackage{makecell}
\usepackage{hyperref}
\usepackage{multirow}
\usepackage{subcaption}
\usepackage{array,multirow,graphicx}
\usepackage{amsmath}
\usepackage{enumitem}
\usepackage[compact]{titlesec}
\usepackage{romannum}

\titlespacing{\paragraph}{1pt}{*0}{*1}

\title{\textsc{Know} How to Make Up Your Mind! Adversarially Detecting and Alleviating Inconsistencies in Natural Language Explanations}

\author{
Myeongjun Jang$^1$~~~
Bodhisattwa Prasad Majumder$^3$~~~
Julian McAuley$^3$~~~ \\
\textbf{Thomas Lukasiewicz}$^{2,1}$~~~
\textbf{Oana-Maria Camburu}$^4$~~~ \\
$^1$University of Oxford, UK 
\texttt{}
$^2$Vienna University of Technology, Austria \\
$^3$University of California San Diego 
\texttt{}
$^4$University College London, UK \\
\texttt{myeongjun.jang@cs.ox.ac.uk}
}

\begin{document}

\maketitle
\begin{abstract}

While recent works have been considerably improving the quality of the natural language explanations (NLEs) generated by a model to justify its predictions, there is very limited research in detecting and alleviating inconsistencies among generated NLEs. In this work, we leverage external knowledge bases to significantly improve on an existing adversarial attack for detecting inconsistent NLEs. We apply our attack to high-performing NLE models and show that models with higher NLE quality do not necessarily generate fewer inconsistencies. Moreover, we propose an off-the-shelf mitigation method to alleviate inconsistencies by grounding the model into external background knowledge. Our method decreases the inconsistencies of previous high-performing NLE models as detected by our attack.

\end{abstract}

\section{Introduction}
The accurate yet black-box nature of deep neural networks 
has accelerated studies on explainable AI. The advent of human-written natural language explanations (NLEs) datasets~\cite{teachmetoexplain} has paved the way for the development of models that provide NLEs for their predictions. However, by introducing an adversarial attack, which we hereafter refer to as eIA (\textbf{e}xplanation \textbf{I}nconsistency \textbf{A}ttack), \citet{camburu2020make} found that an early NLE model \cite{ESNLI} was prone to generate inconsistent NLEs (In-NLEs). More precisely, two \textit{logically contradictory} NLEs generated by a model for two instances that have the same context are considered to form an \textit{inconsistency}. For example, assume a self-driving car stops in a given traffic environment (the context). If the passenger asks the car Q1:``Why did you stop?'', and it provides NLE1: ``Because the traffic light is red.'', and, for the same context, if the passenger instead asks Q2: ``Why did you decide to stop here?'' and the car provides NLE2: ``Because the traffic light is green'', then NLE1 and NLE2 form an inconsistency. 

A model that generates In-NLEs is undesirable, as it \textit{either has a faulty decision-making process} (e.g., the traffic light was green, so the car should not have stopped), or it \textit{generates NLEs that are not faithfully describing its decision-making process} (e.g., the car stopped for a red traffic light, but states that it was green) \cite{camburu2020make}. While recent high-performing NLE models have largely improved in terms of the quality (plausibility) of the generated NLEs, to our knowledge, these models have not been tested against generating inconsistent~NLEs. 

In this work, we first propose a fast, efficient, and task-generalizable adversarial attack that utilizes external knowledge bases. Through experiments on two datasets and four models, we verify the increased efficiency of our approach over the eIA attack, the only inconsistency attack for NLE models, to our knowledge. We also show that the high-performing NLE models are still prone to generating significantly many In-NLEs and, surprisingly, that a higher NLE quality does not necessarily imply fewer inconsistencies. Second, we propose a simple yet efficient off-the-shelf method for alleviating inconsistencies that grounds any NLE model into background knowledge, leading to fewer inconsistencies. The code for this paper is available at \href{https://github.com/MJ-Jang/eKnowIA}{https://github.com/MJ-Jang/eKnowIA}.

\section{Inconsistency Attack}
We propose eKnowIA (\textbf{e}xplanations \textbf{Know}ledge-grounded \textbf{I}nconsistency \textbf{A}ttack), which detects more In-NLEs in a faster and more general manner than eIA.

\subsection{Original eIA Attack}\label{eia}

\paragraph{Setting.}  Given an instance $\mathrm{x}$, \citet{camburu2020make} divide it into: the \textit{context} part $\mathrm{x_c}$ that remains fixed, and the \textit{variable} part $\mathrm{x_v}$ that is changed during the attack. For example, $\mathrm{x_c}$ and $\mathrm{x_v}$ would be a \textit{premise} and a \textit{hypothesis}, respectively, for natural language inference (NLI --- detailed below). Let $\mathrm{e}_m(\mathrm{x})$ denote the NLE generated by a model $m$ for the input $\mathrm{x} = (\mathrm{x_c}, \mathrm{\hat{x}_v})$. The objective is to find $\mathrm{\hat{x}_v}$ such that $\mathrm{e}_m(\mathrm{x})$ and $\mathrm{e}_m((\mathrm{x_c}, \mathrm{\hat{x}_v}))$ are logically contradictory (see examples in Table~\ref{table.inconsist_example}).

\paragraph{Steps.} The eIA attack has the following steps:

\begin{enumerate}[nosep]
    \item Train a neural model to act as a reverse explainer, called R\textsc{ev}E\textsc{xpl}, that takes $\mathrm{x_c}$ and $\mathrm{e}_m(\mathrm{x})$ as input and generates $\mathrm{x_v}$, i.e., R\textsc{ev}E\textsc{xpl}$(\mathrm{x_c};\mathrm{e}_m(\mathrm{x}))=\mathrm{x_v}$.
    
    \item For each generated NLE $\mathrm{e}_m(\mathrm{x})$:
    \begin{enumerate}
        \item \label{step.2a} Automatically create a set of statements $\mathcal{I}_e$ that are inconsistent with $\mathrm{e}_m(\mathrm{x})$.
        \item For each $\hat{e} \in \mathcal{I}_e$, generate a variable part $\mathrm{\hat{x}_v}=$ R\textsc{ev}E\textsc{xpl}$(\mathrm{x_c}; \hat{e})$.
        \item Query $m$ on $\hat{x}=(\mathrm{x_c}, \mathrm{\hat{x}_v})$ to get 
        $\mathrm{e_m}(\hat{\mathrm{x}})$. 
        \item \label{step.2d} Check whether $\mathrm{e_m}(\hat{\mathrm{x}})$ is indeed inconsistent with $\mathrm{e_m}(\mathrm{x})$ by checking whether $\mathrm{e_m}(\hat{\mathrm{x}})$ is included in $\mathcal{I}_e$.
    \end{enumerate}
\end{enumerate}

\paragraph{Creating $\mathcal{I}_e$.}
\citet{camburu2020make} used simple elimination of negation (removing ``not'' or ``n't'') and a task-specific template-based approach for this step. For the template-based approach, they manually create a set of label-specific templates for NLEs such that introducing the instance-specific terms of an NLE from one template into any template from another label creates an inconsistency. They illustrate this process only on the e-SNLI dataset \citep{ESNLI}, leaving room to question how easily it generalizes to other datasets. e-SNLI contains NLEs for the SNLI dataset \cite{SNLI}, where the NLI task consists in identifying whether a \textit{premise} and a \textit{hypothesis} are in a relation of \textit{entailment} (if the premise entails the hypothesis), 
\textit{contradiction} (if the hypothesis contradicts the premise),
or \textit{neutral} (if neither entailment nor contradiction hold). 
Examples of their templates are: ``<X> is <Y>'' (for \textit{entailment}) and ``<X> cannot be <Y>'' (for \textit{contradiction}). Based on the templates,  for a $\mathrm{e}_m(\mathrm{x})$ of ``A dog is an animal.'', an inconsistent statement of ``A dog cannot be an animal'' is obtained (<X> = ``A dog'', <Y> = ``an animal''). They manually identified an average of 10 templates per label. 


\subsection{Our eKnowIA Attack}
The template-based approach in eIA has two major drawbacks: (1) requires substantial human effort to find an exhaustive set of templates for \textit{each} dataset, (2) many different ways of obtaining inconsistencies (e.g., using antonyms) are not taken into account. Moreover, even their negation rule can also be improved. To alleviate these drawbacks, we adopt three rules.

\paragraph{Negation.} We remove and \textit{add} negation tokens to negated and non-negated sentences, respectively. To avoid grammatical errors, we add one negation per sentence only if the sentence belongs to one of the following two templates:

\begin{itemize}[nosep]
    \item <A> is <B>, <A> are <B> (add ``not''),
    \item <A> has <B>, <A> have <B> (add ``does/do not'' only if <B> is a noun).
\end{itemize}

\paragraph{Antonym replacement for adjectives/adverbs.} 
We replace adjectives/adverbs with their antonyms from ConceptNet \cite{ConceptNet} ({using the \href{http://www.nltk.org/api/nltk.tag.html}{NLTK POS tagger}).} Only one adjective or adverb at a time is replaced for each NLE, to avoid deteriorating the contradictory meaning. Employing other abundant thesauruses could improve our approach, which we leave as future work.

\paragraph{Unrelated noun replacement.} We replace a noun with an unrelated one, e.g., ``human'' with ``plant''. This is only applied to the noun that is the last word of the sentence, to reduce the possibility of false inconsistencies as the \ac{POS} tagger occasionally made incorrect predictions for words in the middle of a sentence. To get unrelated nouns, we use the \textit{DistinctFrom} and \textit{Antonym} relations in ConceptNet. However, we noticed that ConceptNet contains noisy triplets where the subject and object are not antonyms, such as ``man'' and ``people'' for ``person''\footnote{A pair of words with \textit{opposite} meanings (from \href{https://en.wikipedia.org/wiki/Opposite\_(semantics)\#Types\_of\_antonyms}{Wikepedia})\hbox{.\!\!}}.
To avoid these,  we created a list (see Table~\ref{table.filtering_triplets} in the appendix) of triplets from ConceptNet to be ignored, by manually investigating a random subset of 3000 detected inconsistencies. While this involved human effort, we highlight that this is due to the nature of ConceptNet and other knowledge bases with more accurate instances may be used instead. However, since we only found eight noisy triplets, we decided to keep ConceptNet, which otherwise worked well for our datasets. Finally, we also noticed that our rules may not lead to In-NLEs if both the context and variable part contain negations. Examples are in Table~\ref{table.filtering_examples} in the appendix. We filter out such pairs.

\begin{table*}[ht!]
	\begin{center}
		\renewcommand{\arraystretch}{1.0}
		\footnotesize{
			\centering{\setlength\tabcolsep{6pt}
		\begin{tabular}{ccccccccc}
		\toprule
		\multirow{2}{*}{Model} & 
		\multicolumn{4}{c}{e-SNLI} & 
		\multicolumn{4}{c}{Cos-E} \\ 
		& Acc. & $\mathcal{S}_r$ & $\mathcal{H}_r$ & e-ViL 
		& Acc.  & $\mathcal{S}_r$ & $\mathcal{H}_r$ & e-ViL
		\\ \hline
	    NILE & 90.7 & 3.13 & 2.27 & 0.80 & - & - & - & -  \\ 
        KnowNILE & \textbf{90.9} &  \textbf{2.42}$\dagger$ & \textbf{1.99}$\dagger$ & \textbf{0.82} & - & - & - & -  \\ \hline

	    CAGE & - & - & - & - &  61.4 & 0.42 & 0.06 & 0.43\\ 
	    KnowCAGE & - & - & - & - &  \textbf{62.6} &  \textbf{0.11}$\dagger$ & \textbf{0.01}$\dagger$ & \textbf{0.44} \\ \hline

	    WT5-base & 90.6 & 12.88 & 1.70 & 0.76 & 65.1   & 0.95 & 0.12 & 0.55 \\
	    KnowWT5-base & \textbf{90.9} &  \textbf{11.45} & \textbf{1.19}$\dagger$ & \textbf{0.80}$\dagger$ & \textbf{65.5} &  \textbf{0.84}$\dagger$ & \textbf{0.09}$\dagger$ & \textbf{0.56} \\ 







		\bottomrule
		\end{tabular}}}
	\vspace*{-1ex}
	\caption{Results of our eKnowIA attack and our method for mitigating In-NLEs. The best results for each pair of (model, Know-model) are in bold; $\mathcal{S}_r$ and $\mathcal{H}_r$ are given in \%; $\dagger$ indicates that Know-models showed statistically significant difference with \textit{p}-value < 0.05 ($\dagger$) using the t-test.}
	\vspace*{-2.5ex}
	\label{table.model_results}%
	\end{center}
\end{table*}

\begin{table}[t!]
	\begin{center}
		\renewcommand{\arraystretch}{1.0}
		\footnotesize{
			\centering{\setlength\tabcolsep{6pt}
		\begin{tabular}{cccccc}
		\toprule

		Dataset &
		Method & 
         Time & $\mathcal{S}_r$ & $\mathcal{H}_r$
		\\ \hline

		\multirow{2}{*}{\makecell{e-SNLI}} 
		& eIA & 10 days & 2.19 & 384/24M\\
		& eKnowIA & \textbf{40 min} & \textbf{12.88} & \textbf{1,494/88K} \\ \hline
		
		\multirow{2}{*}{\makecell{Cos-E}} 
		& eIA & 2.5 days & 0.32 & 5/5M \\
		& eKnowIA & \textbf{5 min} & \textbf{0.95} & \textbf{13/11K} \\
        
		\bottomrule
		\end{tabular}}}
	\vspace*{-1ex}
	\caption{Comparison between eIA and eKnowIA on WT5-base. The best results are in bold; $\mathcal{S}_r$ is given in  \%; $\mathcal{H}_r$ values are in fractions to emphasise the high denominators of the eIA.}
	\vspace*{-2ex}
	\label{table.previous_comparison}%
	\end{center}
	\vspace*{-2ex}
\end{table}

\subsection{Experiments}

\paragraph{Datasets.} We consider two tasks: NLI (with the e-SNLI dataset described in Sec.\ \ref{eia}) and \ac{CQA}. The Cos-E 1.0 dataset \citep{CosE} contains CQA instances formed of a \textit{question}, three \textit{answer candidates}, and an NLE for the correct answer. The objective of the Cos-E \citep{CosE} dataset is to select an \textit{answer} among the three candidates given a \textit{question} and to generate an NLE to support the answer. Following \citet{camburu2020make}, we set the premise as context and the hypothesis as the variable part for e-SNLI. For Cos-E, to avoid omitting the correct answer, we set the question and the correct answer as the context, and the remaining two answer candidates as the variable part. Just like eIA, our attack is solely intended for detecting In-NLEs and not as a label attack (which may or may not happen).

\paragraph{Evaluation metrics.} Let $\mathcal{I}_{e}$ be generated at Step \ref{step.2a} for each instance in a test set $\mathcal{D}_{test}$, and let $\mathcal{I}_s \subseteq \mathcal{I}_{e}$ be the set of detected In-NLEs (after Step \ref{step.2d}). For each instance, our attack can identify multiple inconsistencies (via multiple variable parts). We, therefore, use two evaluation metrics: hit-rate ($\mathcal{H}_r$) and success-rate ($\mathcal{S}_r$):

\vspace{-3.5ex}
\begin{align*}
    \mathcal{S}_r = {N_c}/{|\mathcal{D}_{test}|} \text{ and } 
    \mathcal{H}_r = {|\mathcal{I}_s|}/{|\mathcal{I}_{e}|},
\end{align*}
where $N_c$ is the number of unique instances for which the attack identified at least one inconsistency. Intuitively, $\mathcal{S}_r$ denotes the ratio of the test instances where the attack is successful, while $\mathcal{H}_r$ denotes the ratio of detected In-NLEs to that of the proposed In-NLEs.

\paragraph{Models.} We consider the following high-performing NLE models, with their implementation detailed in Appendix~\ref{sec:appendix_implementation_detail}: NILE~\cite{Nile} for NLI, CAGE~\cite{CosE} for \ac{CQA}, and WT5-base (220M parameters)~\cite{WT5} for both tasks. WT5 models with more parameters (e.g., WT5-11B) would require considerably more computing while providing relatively small gains in NLE quality (32.4 for WT5-base vs.\ 33.7 for WT5-11B \citep{WT5}). Therefore, they are not considered here due to limited computing resources. Implementation details are given in Appendix~\ref{sec:appendix_implementation_detail}. 

\subsection{Results}
\paragraph{eKnowIA vs.\ eIA.}
We compare eKnowIA with eIA only on the WT5-base model, 
since eIA requires a prohibiting amount of time. 
As in \citet{camburu2020make}, we manually verified the naturalness of adversarial hypotheses on 50 random samples for each model. Sentences that go against common sense are considered unnatural. Minor grammatical errors and typos are ignored. We observe that 81.5\% of the adversarial hypotheses were natural, on average, for each model. Details are in Appendix~\ref{section:naturalness}. The results are summarized in Table~\ref{table.previous_comparison}. The e-SNLI results are adjusted to reflect the proportion of natural adversarial hypotheses by multiplying the number of detected pairs of In-NLEs for each model with the estimated naturalness ratio. For Cos-E, an unnatural variable part would consist of stop words or a repetition of another answer candidate. We automatically found 2 out of 22 examples to be unnatural, which were removed. We observe that eIA generates a tremendous amount of inconsistent candidates ($\mathcal{I}_{e}$), e.g., 24M for e-SNLI, thus being extremely slow (e.g., 10 days vs.\ 40 min for eKnowIA), while also obtaining lower $\mathcal{S}_r$ and $\mathcal{H}_r$ than eKnowIA (e.g., 2.19\% vs.\ 12.88\% $\mathcal{S}_r$).

\begin{table*}[ht]
	\begin{center}
		\renewcommand{\arraystretch}{1.0}
		\footnotesize{
			\centering{\setlength\tabcolsep{5.0pt}
   
            \begin{tabular}{p{0.45\linewidth} | p{0.45\linewidth}}
  		\toprule
            \multicolumn{2}{c}{{\makecell{\textsc{Premise}: A man is riding his dirt bike through the air in the desert.
            }}} \\ 
            
            \textsc{Hypothesis}: A man is on a motorbike & \textsc{Hypothesis}: The man is riding a motorbike.  \\
            \textsc{Predicted Label}: entailment & 
            \textsc{Predicted Label}: contradiction     \\
            \textsc{Explanation}: A dirt bike is a motorbike. & 
            \textsc{Explanation}: A dirt bike is not a motorbike. \\  \midrule
            
            \multicolumn{2}{c}{{\makecell{\textsc{Question}: John knew that the sun produced a massive amount of energy in two forms. \\If you were on the surface of the sun, what would kill you first? \\ 
            }}} \\ 

            \textsc{Choices}: heat, light, life on earth & \textsc{Choices}: heat, light, darkness \\
            \textsc{Predicted Label}: heat & 
            \textsc{Predicted Label}: heat \\
            \textsc{Explanation}: the sun produces heat and light. & 
            \textsc{Explanation}: the sun produces heat and darkness. \\
	\bottomrule
	\end{tabular}}}\vspace{-1ex}  
 \caption{Examples of inconsistent NLEs detected by eKnowIA for WT5 on e-SNLI and CAGE on Cos-E. The first column shows the original variable part, and the second column shows the adversarial one.}\label{table.cose_inconsist_example}%
	\end{center}
	\vspace{-0.5ex}
\end{table*}

\begin{table*}[ht]
	\begin{center}
		\renewcommand{\arraystretch}{1.0}
		\footnotesize{
			\centering{\setlength\tabcolsep{5.0pt}
   
            \begin{tabular}{p{0.45\linewidth} | p{0.45\linewidth}}
  		\toprule
            \multicolumn{2}{c}{{\makecell{\textsc{Premise}: A man is riding his dirt bike through the air in the desert. \\ 
            }}} \\ 

            \textsc{Hypothesis}: A man is on a motorbike & \textsc{Hypothesis}: The man is riding a motorbike. \\
            \textsc{Predicted Label}: entailment & 
            \textsc{Predicted Label}: entailment     \\

            \textsc{Extracted Knowledge}: \{dirt bike, IsA, motorcycle\}, \{desert, MannerOf, leave\}, \{air, HasA, oxygen\} &  \textsc{Extracted Knowledge}: \{dirt bike, IsA, motorcycle\}, \{desert, MannerOf, leave\}, \{air, HasA, oxygen\} \\
            \textsc{Explanation}: A dirt bike is a motorbike. & 
            \textsc{Explanation}: A dirt bike is a motorbike. \\ \midrule
            
            \multicolumn{2}{c}{{\makecell{\textsc{Question}: John knew that the sun produced a massive amount of energy in two forms. \\If you were on the surface of the sun, what would kill you first? 
            }}} \\ 

            \textsc{Choices}: heat, light, life on earth & \textsc{Choices}: heat, light, darkness \\
            \textsc{Predicted Label}: heat & 
            \textsc{Predicted Label}: heat \\

            \textsc{Extracted Knowledge}: \{light, IsA, energy\} ,\{heat, IsA, energy\} & \textsc{Extracted Knowledge}: \{light, Antonym, dark\}, \{heat, IsA, energy\} \\
            \textsc{Explanation}: light and heat are two forms of energy. & 
            \textsc{Explanation}: the sun produces heat and light.  \\
	\bottomrule
	\end{tabular}}}\vspace{-1ex}      
 \caption{Examples of successfully defended instances by KnowWT5 on e-SNLI and KnowCAGE on Cos-E. This table should be read together with Table \ref{table.cose_inconsist_example} to appreciate the defence. }\label{table.defended_examples}%
	\end{center}
	\vspace{-1ex}
\end{table*}

\paragraph{eKnowIA on NLE models.}
The results of eKnowIA applied to NILE, CAGE, and WT5 are in the upper lines of each block in Table~\ref{table.model_results}. All models are vulnerable to the inconsistency attack. Also, a better NLE quality may not necessarily guarantee fewer inconsistencies. For example, WT5-base has a better NLE quality than CAGE on Cos-E (0.55 vs.\ 0.43 e-ViL score; see below), but eKnowIA detected more inconsistencies for WT5-base than for CAGE (0.95 vs.\ 0.42 success rate). Examples of generated In-NLEs are in Table~\ref{table.cose_inconsist_example}. More examples are in Tables \ref{table.inconsist_example}--\ref{table.more_inconsist_example} in Appendix~\ref{sec:example_appendix}. We observe that the In-NLEs usually contradict common sense, which is aligned with previous studies showing that language models, used as pre-trained components in the NLE models, often suffer from factual in\-cor\-rectness~\cite{mielke2020linguistic, zhang2021dmrfnet}.

\section{Our \textsc{Know} Method for Alleviating Inconsistencies}
Our approach for alleviating inconsistencies in NLE mo\-dels consists of two steps: (1) extraction of  knowledge related to the input and (2)~knowledge injection.

\paragraph{Extracting related knowledge.}
We leverage a knowledge extraction heuristic proposed by \citet{xu2021fusing} as follows:

\begin{enumerate}[nosep]
    \item Extract entities from an input's context part.
    
    \item Find all knowledge triplets that contain the entities.
    
    \item For each entity, calculate a weight $s_j$ for each extracted triplet as:
    
    \vspace{-3.0ex}
    \begin{align*}
        \mbox{$s_j = w_j \times {N}/{N_{r_j}} \text{ and }
        N = \sum\nolimits_{j=1}^{K} N_{r_j}$,}
    \end{align*}
    %
    where $w_j$ is the weight of the $j$-th triplet pre-defined by the knowledge base (e.g., ConceptNet), $N_{r_j}$ is the number of extracted triplets of the relation $r_j$ for the given instance, and $K$ is the total number of triplets containing the entity for the given instance.
    
    \item For each entity, extract the triplet with the highest score.
\end{enumerate}


\paragraph{Grounding with the extracted knowledge.}
After extracting the triplet with the highest weight per entity in an instance, we transform each of them into natural language and concatenate them to the instance. We use ``Context:'' as a separator between the input and the triplets. We leverage the templates that transform a relation into free-text (e.g., \textit{IsA} to ``is a'') from \citet{petroni2019language}.

\subsection{Experiments}
We apply our \textsc{Know} approach to NILE, CAGE, and WT5-base, and name them KnowNILE, KnowCAGE, and KnowWT5-base, respectively.

\paragraph{Inconsistencies.} The results in Table~\ref{table.model_results} show that grounding in commonsense knowledge diminishes the number of In-NLEs for all models and tasks. The \textsc{Know} models defended against 58\% of the examples attacked by eKnowIA. 
Also, we observed that, among the inconsistent examples of \textsc{Know} models, 20\% of them on average were newly introduced instances. Examples that failed to be defended, as well as newly introduced In-NLEs are provided in Tables~\ref{table.not_defended_example}-\ref{table.newly_introduced} in Appendix~\ref{sec:example_appendix}. Successfully defended examples are provided in Table~\ref{table.defended_examples}. More successfully defended examples, non-defended examples, and newly attacked examples can be found in Tables~\ref{table.attacked_samples_defended}-\ref{table.defended_examples_more} in Appendix~\ref{sec:example_appendix}. 

First, we highlight that a successfully defended example means that our eKNowIE attack did not find an adversarial instance together with which the \textsc{Know} model would form a pair of In-NLEs, while our attack did find at least one such adversarial instance for the original model. Second, we notice that even when the selected knowledge might not be the exact knowledge needed to label an instance correctly, the model can still benefit from this additional knowledge. For example, in the first sample in Table~\ref{table.defended_examples_more} in Appendix~\ref{sec:example_appendix}, the most proper knowledge triplet would be \{dog, DistinctFrom, bird\}. However, despite the indirect knowledge given,\footnote{ConceptNet does not contain \{dog, DistinctFrom, bird\}.} i.e., \{dog, DistinctFrom, cat\}, the model is able to defend the In-NLE by inferring that dogs are different from other animals. 
To examine whether the improved consistency of the \textsc{Know} models stems from \textit{knowledge leakage} (using the same knowledge triplets in the mitigation method as in the attack), we calculate the overlap of triplets. On the e-SNLI dataset, we find that only 0.3\% of knowledge triplets are reused for the attack on the \textsc{Know} models, and no overlap was found for the Cos-E dataset. This indicates that the leakage is not significant.

\paragraph{NLE quality.} To evaluate the quality of generated NLEs, we conducted a human evaluation using Amazon MTurk, as automatic evaluation metrics only weakly reflect human judgements~\cite{evil}.
We follow the setup from \citet{evil}: we asked annotators (three per instance) to judge whether the generated NLEs justify the answer with four options: \{\textit{no}, \textit{weak no}, \textit{weak yes}, \textit{yes}\} and calculated the e-ViL score by mapping them to $\{0, {1}/{3}, {2}/{3}, 1 \}$, respectively. Details of the \mbox{human} evaluation are in Appendix~\ref{sec:human_eval_process}. In Table~\ref{table.model_results}, the \textsc{Know} models show similar NLE quality to their original counterparts, suggesting that our \textsc{Know} method preserves NLE quality while decreasing inconsistencies. Similar results are observed on the automatic evaluation of NLEs (see Appendix~\ref{sec:eval_nles}).

\section{Related Work}
A growing number of works
focus on building NLE models in different areas such as natural language inference~\citep{ESNLI}, question answering~\citep{WT5}, visual-textual reasoning \citep{Hendricks_2018_ECCV, evil, rexc}, medical imaging \citep{mimicnle}, self-driving cars \citep{kim2018textual}, and offensiveness classification~\citep{sap2019social}.
Most commonly, the performance of these models is assessed only in terms of how plausible the reasons provided by their NLEs are. 
To our knowledge, \citet{camburu2020make} is the only work to investigate inconsistencies in NLEs. We improve their adversarial attack as well as bring an approach to alleviate inconsistencies. Works have also been conducted to analyse and make dialogue models generate responses consistent with the dialogue history~\cite{zhang2018personalizing, welleck2019dialogue, li2020dont}. However, these works are difficult to be applied to NLE models, in part because they require specific auxiliary datasets, such as pairs of inconsistent sentences. Other works investigated the logical consistency of a model's predictions~\cite{elazar2021measuring, mitchell2022enhancing, kumar2022striking, lin2022bert}, but would not have straightforward extensions for investigating NLEs inconsistencies.
Besides consistency, NLEs can also be assessed for their faithfulness w.r.t.\ the decision-making process of the model that they aim to explain \citep{wiegreffe2020measuring, pepa}. 

\section{Summary and Outlook}
We proposed the eKnowIA attack, which is more generalizable, successful, and faster than the previous eIA attack in detecting In-NLEs. 
Our experiments show that current NLE models generate a significant number of In-NLEs, and that higher NLE quality does not necessarily imply fewer inconsistencies. We also introduced a simple but efficient method that grounds a model into relevant knowledge, decreasing the number of In-NLEs. Our work paves the way for further work on detecting and alleviating inconsistencies in NLE~models.  

\section*{Limitations}
Our eKnowIA attack contains logical rules designed specifically for the English language. While these rules may apply or be adapted to other languages with simple morphology, there could be languages in which completely new rules may be needed.  
Both our attack and the \textsc{Know} method rely on knowledge bases, which may sometimes be noisy. We employed manual efforts to eliminate (a small number of) noisy triples from ConceptNet. Our attack also relies on a manual annotation to ensure that the adversarial inputs are natural (estimated to be the case 81.5\% of the time). 
Finally, we were not able to test our methods on instances with long text, as we are not aware of datasets with NLEs for long text inputs or long NLEs. 

\section*{Acknowledgements}
This work was partially supported by the Alan Turing Institute under the EPSRC grant
EP/N510129/1, by the AXA Research Fund, and by the EU TAILOR grant 952215. Oana-Maria Camburu was supported by a Leverhulme Early Career Fellowship.
We also acknowledge the use of Oxford’s ARC facility, of the EPSRC-fun\-ded Tier 2 facility JADE~\Romannum{2} (EP/ T022205/1), and of GPU computing support by Scan Computers International Ltd.

\bibliographystyle{acl_natbib}
\bibliography{references}

\begin{thebibliography}{37}
\expandafter\ifx\csname natexlab\endcsname\relax\def\natexlab#1{#1}\fi

\bibitem[{Atanasova et~al.(2023)Atanasova, Camburu, Lioma, Lukasiewicz,
  Simonsen, and Augenstein}]{pepa}
Pepa Atanasova, Oana-Maria Camburu, Christina Lioma, Thomas Lukasiewicz,
  Jakob~Grue Simonsen, and Isabelle Augenstein. 2023.
\newblock {Faithfulness Tests for Natural Language Explanations}.
\newblock In \emph{ACL}.

\bibitem[{Banerjee and Lavie(2005)}]{banerjee2005meteor}
Satanjeev Banerjee and Alon Lavie. 2005.
\newblock {METEOR: A}n automatic metric for {MT} evaluation with improved
  correlation with human judgments.
\newblock In \emph{ACL Workshop on Intrinsic and Extrinsic Evaluation Measures
  for Machine Translation and/or Summarization}.

\bibitem[{Bowman et~al.(2015)Bowman, Angeli, Potts, and Manning}]{SNLI}
Samuel~R. Bowman, Gabor Angeli, Christopher Potts, and Christopher~D. Manning.
  2015.
\newblock \href {https://doi.org/10.18653/v1/D15-1075} {A large annotated
  corpus for learning natural language inference}.
\newblock In \emph{EMNLP}.

\bibitem[{Camburu et~al.(2018)Camburu, Rockt\"{a}schel, Lukasiewicz, and
  Blunsom}]{ESNLI}
Oana-Maria Camburu, Tim Rockt\"{a}schel, Thomas Lukasiewicz, and Phil Blunsom.
  2018.
\newblock \href
  {https://proceedings.neurips.cc/paper/2018/file/4c7a167bb329bd92580a99ce422d6fa6-Paper.pdf}
  {{e-SNLI: N}atural language inference with natural language explanations}.
\newblock In \emph{NeurIPS}, volume~31.

\bibitem[{Camburu et~al.(2020)Camburu, Shillingford, Minervini, Lukasiewicz,
  and Blunsom}]{camburu2020make}
Oana-Maria Camburu, Brendan Shillingford, Pasquale Minervini, Thomas
  Lukasiewicz, and Phil Blunsom. 2020.
\newblock Make up your mind! {A}dversarial generation of inconsistent natural
  language explanations.
\newblock In \emph{ACL}.

\bibitem[{Elazar et~al.(2021)Elazar, Kassner, Ravfogel, Ravichander, Hovy,
  Sch{\"u}tze, and Goldberg}]{elazar2021measuring}
Yanai Elazar, Nora Kassner, Shauli Ravfogel, Abhilasha Ravichander, Eduard
  Hovy, Hinrich Sch{\"u}tze, and Yoav Goldberg. 2021.
\newblock Measuring and improving consistency in pretrained language models.
\newblock \emph{arXiv preprint arXiv:2102.01017}.

\bibitem[{Hendricks et~al.(2018)Hendricks, Hu, Darrell, and
  Akata}]{Hendricks_2018_ECCV}
Lisa~Anne Hendricks, Ronghang Hu, Trevor Darrell, and Zeynep Akata. 2018.
\newblock Grounding visual explanations.
\newblock In \emph{ECCV}.

\bibitem[{Kayser et~al.(2021)Kayser, Camburu, Salewski, Emde, Do, Akata, and
  Lukasiewicz}]{evil}
Maxime Kayser, Oana{-}Maria Camburu, Leonard Salewski, Cornelius Emde, Virginie
  Do, Zeynep Akata, and Thomas Lukasiewicz. 2021.
\newblock {e-ViL: A} dataset and benchmark for natural language explanations in
  vision-language tasks.
\newblock In \emph{ICCV}.

\bibitem[{Kayser et~al.(2022)Kayser, Emde, Camburu, Parsons, Papiez, and
  Lukasiewicz}]{mimicnle}
Maxime Kayser, Cornelius Emde, Oana-Maria Camburu, Guy Parsons, Bartlomiej
  Papiez, and Thomas Lukasiewicz. 2022.
\newblock Explaining chest x-ray pathologies in natural language.
\newblock In \emph{MICCAI}, pages 701--713.

\bibitem[{Kim et~al.(2018)Kim, Rohrbach, Darrell, Canny, and
  Akata}]{kim2018textual}
Jinkyu Kim, Anna Rohrbach, Trevor Darrell, John Canny, and Zeynep Akata. 2018.
\newblock Textual explanations for self-driving vehicles.
\newblock In \emph{ECCV}.

\bibitem[{Kumar and Joshi(2022)}]{kumar2022striking}
Ashutosh Kumar and Aditya Joshi. 2022.
\newblock \href {https://doi.org/10.18653/v1/2022.findings-acl.148} {Striking a
  balance: Alleviating inconsistency in pre-trained models for symmetric
  classification tasks}.
\newblock In \emph{Findings of ACL}, pages 1887--1895.

\bibitem[{Kumar and Talukdar(2020)}]{Nile}
Sawan Kumar and Partha Talukdar. 2020.
\newblock \href {https://doi.org/10.18653/v1/2020.acl-main.771} {{NILE}:
  Natural language inference with faithful natural language explanations}.
\newblock In \emph{ACL}.

\bibitem[{Li et~al.(2020)Li, Roller, Kulikov, Welleck, Boureau, Cho, and
  Weston}]{li2020dont}
Margaret Li, Stephen Roller, Ilia Kulikov, Sean Welleck, Y-Lan Boureau,
  Kyunghyun Cho, and Jason Weston. 2020.
\newblock \href {https://doi.org/10.18653/v1/2020.acl-main.428} {Don{'}t say
  that! {M}aking inconsistent dialogue unlikely with unlikelihood training}.
\newblock In \emph{ACL}.

\bibitem[{Lin(2004)}]{ROUGE}
Chin-Yew Lin. 2004.
\newblock {ROUGE: A} package for automatic evaluation of summaries.
\newblock In \emph{Text Summarization Branches Out}.

\bibitem[{Lin and Ng(2022)}]{lin2022bert}
Ruixi Lin and Hwee~Tou Ng. 2022.
\newblock \href {https://doi.org/10.18653/v1/2022.acl-short.11} {Does {BERT}
  know that the {IS}-a relation is transitive?}
\newblock In \emph{ACL}, pages 94--99.

\bibitem[{Loshchilov and Hutter(2018)}]{AdamW}
Ilya Loshchilov and Frank Hutter. 2018.
\newblock Decoupled weight decay regularization.
\newblock In \emph{ICLR}.

\bibitem[{Majumder et~al.(2022)Majumder, Camburu, Lukasiewicz, and
  McAuley}]{rexc}
Bodhisattwa~Prasad Majumder, Oana-Maria Camburu, Thomas Lukasiewicz, and Julian
  McAuley. 2022.
\newblock Rationale-inspired natural language explanations with commonsense.
\newblock In \emph{ICML}.

\bibitem[{Marasovi{\'c} et~al.(2020)Marasovi{\'c}, Bhagavatula, Park, Le~Bras,
  Smith, and Choi}]{marasovic2020natural}
Ana Marasovi{\'c}, Chandra Bhagavatula, Jae~sung Park, Ronan Le~Bras, Noah~A.
  Smith, and Yejin Choi. 2020.
\newblock \href {https://doi.org/10.18653/v1/2020.findings-emnlp.253} {Natural
  language rationales with full-stack visual reasoning: From pixels to semantic
  frames to commonsense graphs}.
\newblock In \emph{Findings of EMNLP}.

\bibitem[{Marasović et~al.(2022)Marasović, Beltagy, Downey, and
  Peters}]{marasovi2021fewshot}
Ana Marasović, Iz~Beltagy, Doug Downey, and Matthew~E. Peters. 2022.
\newblock Few-shot self-rationalization with natural language prompts.
\newblock In \emph{Findings of NAACL}.

\bibitem[{Mielke et~al.(2020)Mielke, Szlam, Boureau, and
  Dinan}]{mielke2020linguistic}
Sabrina~J. Mielke, Arthur Szlam, Y-Lan Boureau, and Emily Dinan. 2020.
\newblock Linguistic calibration through metacognition: {A}ligning dialogue
  agent responses with expected correctness.
\newblock \emph{arXiv preprint arXiv:2012.14983}.

\bibitem[{Mitchell et~al.(2022)Mitchell, Noh, Li, Armstrong, Agarwal, Liu,
  Finn, and Manning}]{mitchell2022enhancing}
Eric Mitchell, Joseph~J. Noh, Siyan Li, William~S. Armstrong, Ananth Agarwal,
  Patrick Liu, Chelsea Finn, and Christopher~D. Manning. 2022.
\newblock Enhancing self-consistency and performance of pre-trained language
  models through natural language inference.
\newblock \emph{arXiv preprint arXiv:2211.11875}.

\bibitem[{Narang et~al.(2020)Narang, Raffel, Lee, Roberts, Fiedel, and
  Malkan}]{WT5}
Sharan Narang, Colin Raffel, Katherine Lee, Adam Roberts, Noah Fiedel, and
  Karishma Malkan. 2020.
\newblock {WT5?! T}raining text-to-text models to explain their predictions.
\newblock \emph{arXiv preprint arXiv:2004.14546}.

\bibitem[{Papineni et~al.(2002)Papineni, Roukos, Ward, and Zhu}]{BLEU}
Kishore Papineni, Salim Roukos, Todd Ward, and Wei-Jing Zhu. 2002.
\newblock {BLEU: A} method for automatic evaluation of machine translation.
\newblock In \emph{ACL}.

\bibitem[{Petroni et~al.(2019)Petroni, Rockt{\"a}schel, Riedel, Lewis, Bakhtin,
  Wu, and Miller}]{petroni2019language}
Fabio Petroni, Tim Rockt{\"a}schel, Sebastian Riedel, Patrick Lewis, Anton
  Bakhtin, Yuxiang Wu, and Alexander Miller. 2019.
\newblock Language models as knowledge bases?
\newblock In \emph{EMNLP-IJCNLP}.

\bibitem[{Raffel et~al.(2020)Raffel, Shazeer, Roberts, Lee, Narang, Matena,
  Zhou, Li, and Liu}]{T5}
Colin Raffel, Noam Shazeer, Adam Roberts, Katherine Lee, Sharan Narang, Michael
  Matena, Yanqi Zhou, Wei Li, and Peter~J. Liu. 2020.
\newblock \href {https://w.jmlr.org/papers/volume21/20-074/20-074.pdf}
  {Exploring the limits of transfer learning with a unified text-to-text
  transformer}.
\newblock \emph{JMLR}, 21.

\bibitem[{Rajani et~al.(2019)Rajani, McCann, Xiong, and Socher}]{CosE}
Nazneen~Fatema Rajani, Bryan McCann, Caiming Xiong, and Richard Socher. 2019.
\newblock \href {https://doi.org/10.18653/v1/P19-1487} {Explain yourself!
  {L}everaging language models for commonsense reasoning}.
\newblock In \emph{ACL}.

\bibitem[{Rouse(2019)}]{rouse2019reliability}
Steven~V. Rouse. 2019.
\newblock Reliability of {MTurk} data from masters and workers.
\newblock \emph{Journal of Individual Differences}.

\bibitem[{Sap et~al.(2019)Sap, Gabriel, Qin, Jurafsky, Smith, and
  Choi}]{sap2019social}
Maarten Sap, Saadia Gabriel, Lianhui Qin, Dan Jurafsky, Noah~A. Smith, and
  Yejin Choi. 2019.
\newblock Social bias frames: Reasoning about social and power implications of
  language.
\newblock \emph{arXiv preprint arXiv:1911.03891}.

\bibitem[{Speer et~al.(2017)Speer, Chin, and Havasi}]{ConceptNet}
Robyn Speer, Joshua Chin, and Catherine Havasi. 2017.
\newblock {ConceptNet 5.5: A}n open multilingual graph of general knowledge.
\newblock In \emph{AAAI}.

\bibitem[{Welleck et~al.(2019)Welleck, Weston, Szlam, and
  Cho}]{welleck2019dialogue}
Sean Welleck, Jason Weston, Arthur Szlam, and Kyunghyun Cho. 2019.
\newblock \href {https://doi.org/10.18653/v1/P19-1363} {Dialogue natural
  language inference}.
\newblock In \emph{ACL}.

\bibitem[{Wiegreffe and Marasovic(2021)}]{teachmetoexplain}
Sarah Wiegreffe and Ana Marasovic. 2021.
\newblock Teach me to explain: {A} review of datasets for explainable {NLP}.
\newblock In \emph{NeurIPS}, volume~35.

\bibitem[{Wiegreffe et~al.(2021)Wiegreffe, Marasovic, and
  Smith}]{wiegreffe2020measuring}
Sarah Wiegreffe, Ana Marasovic, and Noah~A. Smith. 2021.
\newblock Measuring association between labels and free-text rationales.
\newblock In \emph{EMNLP}.

\bibitem[{Xu et~al.(2021)Xu, Zhu, Xu, Liu, Zeng, and Huang}]{xu2021fusing}
Yichong Xu, Chenguang Zhu, Ruochen Xu, Yang Liu, Michael Zeng, and Xuedong
  Huang. 2021.
\newblock \href {https://doi.org/10.18653/v1/2021.findings-acl.102} {Fusing
  context into knowledge graph for commonsense question answering}.
\newblock In \emph{Findings of ACL}.

\bibitem[{Yordanov et~al.(2022)Yordanov, Kocijan, Lukasiewicz, and
  Camburu}]{yordan2021fewshot}
Yordan Yordanov, Vid Kocijan, Thomas Lukasiewicz, and Oana-Maria Camburu. 2022.
\newblock {Few-Shot Out-of-Domain Transfer of Natural Language Explanations}.
\newblock In \emph{Findings of EMNLP}.

\bibitem[{Zhang et~al.(2018)Zhang, Dinan, Urbanek, Szlam, Kiela, and
  Weston}]{zhang2018personalizing}
Saizheng Zhang, Emily Dinan, Jack Urbanek, Arthur Szlam, Douwe Kiela, and Jason
  Weston. 2018.
\newblock \href {https://doi.org/10.18653/v1/P18-1205} {Personalizing dialogue
  agents: {I} have a dog, do you have pets too?}
\newblock In \emph{ACL}.

\bibitem[{Zhang et~al.(2020)Zhang, Kishore, Wu, Weinberger, and
  Artzi}]{zhang2019bertscore}
Tianyi Zhang, Varsha Kishore, Felix Wu, Kilian~Q. Weinberger, and Yoav Artzi.
  2020.
\newblock {BERTScore: E}valuating text generation with {BERT}.
\newblock In \emph{ICLR}.

\bibitem[{Zhang et~al.(2021)Zhang, Yu, Zhao, and Ran}]{zhang2021dmrfnet}
Weifeng Zhang, Jing Yu, Wenhong Zhao, and Chuan Ran. 2021.
\newblock {DMRFNet: D}eep multimodal reasoning and fusion for visual question
  answering and explanation generation.
\newblock \emph{Information Fusion}, 72:70--79.

\end{thebibliography}

\begin{acronym}
    \acro{DNN}{deep neural network}
    \acro{NLP}{natural language processing}
    \acro{PLM}{pre-trained language model}
    \acro{POS}{part-of-speech}
    \acro{NLI}{natural language inference}
    \acro{CQA}{commonsense question answering}
    \acro{AMT}{Amazon Mechanical Turk}
\end{acronym}

\clearpage
\appendix

\section{Appendix} \label{sec:appendix}

\subsection{Implementation Details}\label{sec:appendix_implementation_detail}
We implemented the WT5-base model based on the HuggingFace transformers package\footnote{\href{https://github.com/huggingface/transformers}{https://github.com/huggingface/transformers}} and replicated performance close to the reported results (see  Section~\ref{sec:wt5_perf}). For the other models, we used the implementations provided by the respective authors. A single Titan X GPU was used. 

\subsection{Training R\textsc{ev}E\textsc{xpl}} \label{sec:appdneix_revexpl} 

We adopted T5-base~\cite{T5} for training the reverse explainer (R\textsc{ev}E\textsc{xpl}). We trained the model for 30 epochs with a batch size of 8. For efficient training, early stopping was applied if the validation loss increases for 10 consecutive logging steps, which were set to 30,000 iterations. The dropout ratio was set to 0.1. We used the AdamW optimiser \cite{AdamW} with learning rate $1e^{-4}$ and epsilon $1e^{-8}$. We also used gradient clipping to a maximum norm of 1.0 and a linear learning rate schedule decaying from $5e^{-5}$.

For Cos-E, we used 10\% of the training data as the validation set, and the original validation set as the test set. 

\subsection{WT5-base Performance Replication} \label{sec:wt5_perf}
This section describes the performance of our trained WT5-base model. We report the accuracy for measuring the performance on the \ac{NLI} and \ac{CQA} tasks. To automatically evaluate the quality of generated NLEs, we use the BLEU score \cite{BLEU}, ROUGE  \cite{ROUGE}, Meteor ~\cite{banerjee2005meteor}, and the BERT score~\cite{zhang2019bertscore}, which are widely used automatic evaluation metrics. The results are summarised in Table \ref{table.WT5_res}. In terms of accuracy and BLEU score, our replication performs better than originally reported for Cos-E, but produced slightly lower results for e-SNLI.

\begin{table}[ht]
	\begin{center}
		\renewcommand{\arraystretch}{1.1}
		\footnotesize{
			\centering{\setlength\tabcolsep{1.0pt}
				\begin{tabular}{ccccccccc}
					\toprule
					 & & Acc. & BLEU & R-1 & R-2 & R-L & Meteor & BERT-S \\
					 \hline
					 \multirow{2}{*}{\makecell{e-SNLI}} & ours 
                     & 90.6 & 28.4 & 45.8 & 22.5 & 40.6 & 33.7 & 89.8  \\ 
                     
					 & reported 
                     & 90.9 & 32.4 & - & - & - & - & -  \\ 
                     \hline
                     
					 \multirow{2}{*}{\makecell{Cos-E}} & ours
                     & 65.3 & 7.3 & 25.0 & 8.3 & 21.6 & 20.2 & 86.3  \\ 
                     & reported & 59.4 & 4.6 & - & - & - & - & -  \\ 
					\bottomrule	
		\end{tabular}}}
	\caption{Performance of our implementation of WT5-base on e-SNLI and Cos-E. The notations R-1, R-2, R-L, and BERT-S denote ROUGE-1, ROUGE-2, ROUGE-L score, and BERT-Score, respectively. } \label{table.WT5_res}%
	\end{center}
	\vspace{-2ex}
\end{table}

\subsection{Naturalness Evaluation of the Generated Variable Parts}\label{section:naturalness}
It could be unfair to consider that a model generates inconsistent NLEs if the adversarial variable parts are unnatural. Hence, we manually evaluated 50 random samples of generated adversarial variable parts for each model (or all samples when there were less than 50 pairs of inconsistencies found).  

On e-SNLI, we observe that, on average, 81.5\% ($\pm$ 1.91) of the reverse variable parts were natural instances, i.e., semantically valid and not contradicting commonsense. The specific figures for each e-SNLI model were 80\%, 80\%, 84\%, and 82\% for KnowNILE, NILE, WT5, and KnowWT5, respectively. We adapted the results in Table \ref{table.model_results} to reflect the number of inconsistencies caused only by natural variable parts.

For the Cos-E dataset, we considered that the variable parts (the two incorrect answer choices) are unnatural if (1) the answer choices are stopwords of the NLTK package or (2) the correct answer is repeated. We observed only one unnatural case for KnowWT5 and WT5, respectively, and none for the other two models. We eliminated the two cases from the counts.

\subsection{Design of Human Evaluation Process for Assessing NLE Quality}\label{sec:human_eval_process}
For the human evaluation, we sampled 200 generated NLEs for each model. 
Three Anglophone annotators are employed per instance.  We selected annotators with a Lifetime HITs acceptance rate of at least 98\% and an accepted number of HITs greater than 1,000.
However, it is widely known that the quality of MTurk annotation is not guaranteed even for Master workers~\cite{rouse2019reliability}. When we used the e-ViL evaluation framework off-the-shelf \citep{evil}, we found that many workers do annotations without due consideration by simply checking ``\textit{yes}'' in most cases. We also initially obtained an inter-annotator agreement captured by Fleiss's Kappa ($\mathcal{K}$) of only 0.06 on average for  Cos-E, which casted doubt on the quality of the evaluation. 
This prompted us to add a quality control measure to the evaluation framework. 
We carefully collected \textit{trusted examples} where the quality of the NLEs is objectively ``\textit{yes}'' or ``\textit{no}''. For each HIT consisting of 10 examples, we incorporated in random locations two trusted examples with the correct answers being ``\textit{yes}'' and ``\textit{no}'', respectively. After annotation, we discarded the HITs where the annotators gave a wrong answer for any of the trusted examples (we consider correct a ``weak yes'' answer for a ``yes'' trusted example and a ``weak no'' for a ``no'' trusted example). We repeated this process until the number of rejected HITs was fewer than 15\% of the total HITs. We achieved an increased $\mathcal{K}$ value of 0.46 and 0.34 for e-SNLI and Cos-E, respectively, from 0.35 and 0.06 (without trusted~examples). Similar levels of $\mathcal{K}$ as ours were obtained in other studies, such as \cite{marasovi2021fewshot, yordan2021fewshot}.

\subsection{Quality Evaluation on the Generated NLEs}\label{sec:eval_nles}
Table~\ref{table.human_eval_detail_result} shows the detailed results of human evaluation on the quality of generated NLEs. In addition to the e-ViL score, we followed the evaluation method of \citet{marasovic2020natural} by merging \textit{weak no} and \textit{weak yes} to \textit{no} and \textit{yes}, respectively, and reporting the ratios of \textit{w/yes} and \textit{w/no}. Also, the results of the automatic evaluation metrics are provided in Table~\ref{table.automatic_performance}. The results show that all the Know-models show similar or better results than their original counterparts.

\begin{table}[t!]
	\begin{center}
		\renewcommand{\arraystretch}{1.0}
		\footnotesize{
			\centering{\setlength\tabcolsep{1.0pt}
		\begin{tabular}{c|ccc|ccc}
		\toprule
        \multirow{2}{*}{Model} & \multicolumn{3}{c|}{e-SNLI} & \multicolumn{3}{c}{Cos-E} \\
	    & e-ViL & W/Yes & W/No  
	    & e-ViL & W/Yes & W/No   \\ \hline
	    CAGE & - & - & - & 0.43 & 46 & 54 \\
	    KnowCAGE & - & - & - & \textbf{0.44} & \textbf{47} & \textbf{53} \\ \hline
	    
	    NILE & 0.80 & 83 & 17  & - & - & - \\
	    KnowNILE & \textbf{0.82} & \textbf{86} & \textbf{14}  & - & - & - \\ \hline
	    
	    WT5 & 0.76 & 80 & 20  & 0.55 & 55 & 45  \\
	    KnowWT5 & \textbf{0.80} & \textbf{84} & \textbf{16} & \textbf{0.56} & \textbf{57} & \textbf{43}  \\
	    
		\bottomrule
		\end{tabular}}}
	\vspace*{-1ex}
	\caption{Human evaluation results on the generated NLEs. The number of ``W/Yes'' (merged ``Yes'' and ``Weak Yes'') and ``W/No'' (``No'' and ``Weak No'') are in \%. The best results are in bold.}
	\vspace*{-1ex}
	\label{table.human_eval_detail_result}%
	\end{center}
	\vspace*{-1ex}
\end{table}

\begin{table}[t!]
	\begin{center}
		\renewcommand{\arraystretch}{1.1}
		\footnotesize{
			\centering{\setlength\tabcolsep{1.0pt}
				\begin{tabular}{c|ccccccc}
					\toprule
					 & & BLEU & R-1 & R-2 & R-L & Meteor & BERT-S \\
					 \hline
					 \multirow{4}{*}{\makecell{e-SNLI}}  & WT5-base 
                     &  28.4 & 45.8 & 22.5 & 40.6 & 33.7 & 89.8  \\ 
					 & KnowWT5-base 
                     &  \textbf{30.6} & \textbf{48.2} & \textbf{24.6} & \textbf{43.0} & \textbf{38.0} & \textbf{90.5}  \\
                     & NILE 
                     & 22.3 & 41.7 & 18.7 & 36.3 & 30.2 & 90.0  \\ 
					 & KnowNILE 
                     & \textbf{22.4} & \textbf{42.0} & \textbf{18.9} & \textbf{36.5} & \textbf{30.5} & \textbf{90.1}  \\ 
                     \hline
					 \multirow{4}{*}{\makecell{Cos-E}} & WT5-base
                     & 7.3 & 25.0 & 8.3 & 21.6 & 20.2 & 86.3  \\ 
                     & KnowWT5-base & \textbf{7.9} & \textbf{26.7} & \textbf{9.6} & \textbf{22.9} & \textbf{21.8} & \textbf{86.7}  \\ 
                     & CAGE
                     & 3.0 & 9.7 & \textbf{1.1} & 9.0 & 6.3 & 85.1  \\ 
                     & KnowCAGE & 3.0 & \textbf{9.8} & 1.0 & 9.0 & \textbf{6.4} & 85.1  \\ 
					\bottomrule	
		\end{tabular}}}
	\caption{Automatic evaluation results on generated NLEs on the e-SNLI and Cos-E datasets. The notations R-1, R-2, R-L, and BERT-S denote ROUGE-1, ROUGE-2, ROUGE-L score, and BERT-Score, respectively. The best results are in bold.} \label{table.automatic_performance}%
	\end{center}
	\vspace{-3ex}
\end{table}

\begin{table}[t]
	\begin{center}
		\renewcommand{\arraystretch}{1.5}
		\footnotesize{
			\centering{\setlength\tabcolsep{2.0pt}
				\begin{tabular}{c|c|c}
					\toprule
					Subject & Relation & Object \\ \hline
					men & Antonym & humans \\
					man & Antonym & person \\
					woman & Antonym & person \\
					people & Antonym & person \\
					flower & DistinctFrom & plant \\
					politician & Antonym & man \\
					children & Antonym & people \\
					\bottomrule	
		\end{tabular}}}
	\vspace{-1ex}
	\caption{List of filtered noisy triplets in ConceptNet.} \label{table.filtering_triplets}%
	\end{center}
	\vspace{-3.5ex}
\end{table}

\onecolumn
\subsection{Examples}\label{sec:example_appendix}

\begin{table*}[ht]
	\begin{center}
		\renewcommand{\arraystretch}{1.0}
		\footnotesize{
			\centering{\setlength\tabcolsep{5pt}
				\begin{tabular}{l|l}
					\toprule
					\makecell[c]{\textsc{Original} \textsc{Explanation}} & \makecell[c]{\textsc{Reverse} \textsc{Explanation}} \\
					\midrule
					Not all men are teaching science. & Not all men are teaching biology. \\
					A dog is not a car. & A dog is not a bike. \\
					The boy is not necessarily looking at another boy. & The boy is not necessarily looking at another female. \\
					A child is not a man. & A child is not a wife. \\
					A bird is not a squirrel. & A bird is not a moose. \\
					A group of dogs is not a woman. & A group of dogs is not a person. \\
					\bottomrule	
		\end{tabular}}}
	\caption{Examples where both the original and reverse NLEs contain negation expressions. These NLEs are not contradictory with each other.} \label{table.filtering_examples}%
	\end{center}
\end{table*}

\begin{table*}[ht]
	\begin{center}
		\renewcommand{\arraystretch}{1.0}
		\footnotesize{
			\centering{\setlength\tabcolsep{5pt}
				\begin{tabular}{l|l}
					\toprule
     
                    \multicolumn{2}{c}{\makecell{\textsc{Premise}: Two hussars sit perched on horses, dressed in extravagant ceremonial wear, each holding a sabre in their \\ right hand, reigns to the horse in their left.}}                \\

                    \textsc{Hypothesis}: There are professional riders at a ceremony. &
                    \textsc{Hypothesis}: Two amateur riders are riding horses.\\
                    \textsc{Predicted} \textsc{Label}: Entailment & 
                    \textsc{Predicted} \textsc{Label}: Entailment       \\
                    \textsc{Explanation}: Hussars are professional riders. & 
                    \textsc{Explanation}: Hussars are amateur riders. \\
                    \midrule

                    \multicolumn{2}{c}{\textsc{Premise}: A cheerleader in a tight red and white uniform is passing out white t-shirts at a sporting event.}                \\

                    \textsc{Hypothesis}: A player passes out hotdogs. &
                    \textsc{Hypothesis}: A player is passing out shirts.\\
                    \textsc{Predicted} \textsc{Label}: Contradiction & 
                    \textsc{Predicted} \textsc{Label}: Entailment       \\
                    \textsc{Explanation}: A cheerleader is not a player. & 
                    \textsc{Explanation}: A cheerleader is a player. \\
                    \midrule
                    
                    \multicolumn{2}{c}{\textsc{Premise}: Two people using a water buffalo to cultivate a watery field.}                \\

                    \textsc{Hypothesis}: Two people are outside with animals. &
                    \textsc{Hypothesis}: Two people are using a plant.\\
                    \textsc{Predicted} \textsc{Label}: Entailment & 
                    \textsc{Predicted} \textsc{Label}: Entailment       \\
                    \textsc{Explanation}: A water buffalo is an animal. & 
                    \textsc{Explanation}: A water buffalo is a plant. \\
                    \midrule

                    \multicolumn{2}{c}{\makecell{\textsc{Question}: Crabs live in what sort of environment?}}                \\

                    \textsc{Choices}: bodies of water, saltwater, galapagos &
                    \textsc{Choices}: bodies of earth, saltwater, atlantic ocean \\
                    \textsc{Predicted} \textsc{Answer}: bodies of water  & 
                    \textsc{Predicted} \textsc{Answer}: bodies of earth       \\
                    
                    \makecell[l]{\textsc{Explanation}: Crabs live in bodies of water.} & 
                    \makecell[l]{\textsc{Explanation}: Crabs live in bodies of earth.} \\
                    \midrule
                    
                    \multicolumn{2}{c}{\makecell{\textsc{Question}: The piece of paper was worth a lot of money, it was an old Apple Inc what?}} \\

                    \textsc{Choices}: stock certificate, copy machine, ream &
                    \textsc{Choices}: stock certificate, piece of stone, book\\
                    \textsc{Predicted} \textsc{Answer}: stock certificate & 
                    \textsc{Predicted} \textsc{Answer}: stock certificate       \\
                    
                    \makecell[l]{\textsc{Explanation}: A stock certificate is the only thing \\ that is not a piece of paper.} & 
                    \makecell[l]{\textsc{Explanation}: A stock certificate is the only thing \\ that is a piece of paper.} \\ \midrule

                    \multicolumn{2}{c}{\makecell{\textsc{Question}: When a person admits his mistakes, what are they doing?}} \\

                    \textsc{Choices}: act responsibly, learn to swim, feel relieved &
                    \textsc{Choices}: act responsibly, think critically, act irresponsibly\\
                    \textsc{Predicted} \textsc{Answer}: act responsibly & 
                    \textsc{Predicted} \textsc{Answer}: act irresponsibly      \\
                    
                    \makecell[l]{\textsc{Explanation}: when a person admits his mistakes, \\ they act responsibly.} & 
                    \makecell[l]{\textsc{Explanation}: when a person admits his mistakes, \\ they act irresponsibly.} \\                    
                    
					\bottomrule	
		\end{tabular}}}
	\caption{Examples of inconsistent NLEs detected by eKnowIA for the WT5-base model on e-SNLI and Cos-E. The first column shows the original variable part and the second column shows the adversarial one.} \label{table.inconsist_example}%
	\end{center}
	\vspace{-2.5ex}
\end{table*}

\begin{table*}[ht]
	\begin{center}
		\renewcommand{\arraystretch}{1.0}
		\footnotesize{
			\centering{\setlength\tabcolsep{5pt}
				\begin{tabular}{l|l}
					\toprule
                    \multicolumn{2}{c}{\textsc{Premise}: These are young adults who seem to be working together to protect the plants surrounding the white pole.}                \\
                    \textsc{Hypothesis}: The adults are young. &
                    \textsc{Hypothesis}: The young adults are underdeveloped.\\
                    \textsc{Predicted} \textsc{Label}: Entailment & 
                    \textsc{Predicted} \textsc{Label}: Entailment       \\
                    \textsc{Explanation}: Young adults are young. & 
                    \textsc{Explanation}: Young adults are seniors. \\
                    \midrule
                    
                    \multicolumn{2}{c}{\makecell{\textsc{Premise}: A young boy with close-cropped hair, wearing a red robe, is holding a black kettle as someone is about to \\ pour something in it.}}                \\
                    \makecell[l]{\textsc{Hypothesis}: a small boy is holding a kettle that will \\ soon be filled} &
                    \textsc{Hypothesis}: The small boy is not underdeveloped.\\
                    \textsc{Predicted} \textsc{Label}: Entailment & 
                    \textsc{Predicted} \textsc{Label}: Entailment       \\
                    \makecell[l]{\textsc{Explanation}: A young boy is a small boy.} & 
                    \makecell[l]{\textsc{Explanation}: A young boy is a small female.} \\
                    \midrule
                    
                    \multicolumn{2}{c}{\textsc{Premise}: A group of people in a room, most with computers, some raising their hands.}                \\
                    \textsc{Hypothesis}: Some people have questions. &
                    \textsc{Hypothesis}: A group of people are asking questions.\\
                    \textsc{Predicted} \textsc{Label}: Neutral & 
                    \textsc{Predicted} \textsc{Label}: Contradiction       \\
                    \makecell[l]{\textsc{Explanation}: People raising their hands are not \\ asking questions.} & 
                    \makecell[l]{\textsc{Explanation}: People raising their hands are asking \\ questions.} \\
					\bottomrule	
		\end{tabular}}}
	\vspace{-1ex}
	\caption{Examples of inconsistent NLEs detected by our eKnowIA attack for the NILE model on e-SNLI. The first column shows the original hypothesis, and the second one shows the adversarial hypothesis from our attack.} \label{table.nile_inconsist_example}%
	\end{center}
	\vspace{-0.5ex}
\end{table*}

\begin{table*}[ht]
	\begin{center}
		\renewcommand{\arraystretch}{1.0}
		\footnotesize{
			\centering{\setlength\tabcolsep{3pt}
				\begin{tabular}{l|l}
					\toprule
                    \multicolumn{2}{c}{\makecell{\textsc{Question}: A good interview after applying for a job may cause you to feel what?}} \\
                    
                    \textsc{Choices}: hope, income, offer &
                    \textsc{Choices}: hope, resentment, fear \\
                    \textsc{Predicted} \textsc{Answer}: hope & 
                    \textsc{Reverse} \textsc{Answer}: hope     \\
                    
                    \makecell[l]{\textsc{Explanation}: hope is the only thing that would cause you \\ to feel hope.} & 
                    \makecell[l]{\textsc{Explanation}: hope is the only thing that would cause you \\ to feel fear.} \\
                    \midrule
                    
                    \multicolumn{2}{c}{\makecell{\textsc{Question}: What does a stove do to the place that it's in?}}\\

                    \textsc{Choices}: warm room, brown meat, gas or electric &
                    \textsc{Choices}: warm room, cook food, heat the outside \\
                    \textsc{Predicted} \textsc{Answer}: warm room & 
                    \textsc{Reverse} \textsc{Answer}: heat the outside       \\
                    
                    \makecell[l]{\textsc{Explanation}: a stove heats the room.} & 
                    \makecell[l]{\textsc{Explanation}: a stove heats the outside.} \\
                    \midrule
		\end{tabular}}}
	\vspace{-1.5ex}
	\caption{Examples of inconsistent NLEs detected by our eKnowIA attack for the CAGE model on  Cos-E. The first column shows the original hypothesis, and the second column shows  the adversarial hypothesis from our attack.} \label{table.more_inconsist_example}%
	\end{center}
    \vspace{-0.5ex}
\end{table*}

\begin{table*}[ht]
	\begin{center}
		\renewcommand{\arraystretch}{1.0}
		\footnotesize{
			\centering{\setlength\tabcolsep{5.0pt}
   
            \begin{tabular}{p{0.45\linewidth} | p{0.45\linewidth}}
  		\toprule
            \multicolumn{2}{c}{\makecell[c]{\textsc{Premise}: A dog standing near snow looking at water.
            }} \\
            
            \textsc{Hypothesis}: A bird is standing near snow.  & \textsc{Hypothesis}: A bird is near water.  \\
            \textsc{Predicted Label}: contradiction & 
            \textsc{Predicted Label}: entailment \\
            \textsc{Explanation}: A dog is not a bird. & 
            \textsc{Explanation}: A dog is a bird. \\  \midrule
            
            \multicolumn{2}{c}{{\makecell{\textsc{Question}: What is a person who is good at sports considered?
            }}} \\ 
            \textsc{Choices}: talented, affluent, reproduce & \textsc{Choices}: talented, untalented, good at \\
            \textsc{Predicted Label}: talented & 
            \textsc{Predicted Label}: untalented \\
            \textsc{Explanation}: a person who is good at sports is considered talented. & 
            \textsc{Explanation}: a person who is good at sports is considered untalented \\
	\bottomrule
	\end{tabular}}}\vspace{-1ex}
 \caption{Examples of inconsistent NLEs detected by eKnowIA for  NILE on e-SNLI and WT5 on Cos-E. The first column shows the original variable part, and the second column shows the adversarial one.}\label{table.attacked_samples_defended}%
	\end{center}
	\vspace{-0.5ex}
\end{table*}

\begin{table*}[ht]
	\begin{center}
		\renewcommand{\arraystretch}{1.0}
		\footnotesize{
			\centering{\setlength\tabcolsep{5.0pt}
   
            \begin{tabular}{p{0.45\linewidth} | p{0.45\linewidth}}
  		\toprule
            \multicolumn{2}{c}{\makecell[c]{\textsc{Premise}: A dog standing near snow looking at water. \\ \textsc{Hypothesis}: A bird is near water. \\ 
            }} \\

            \textsc{Hypothesis}: A bird is standing near snow. & \textsc{Hypothesis}: A bird is near water. \\
            \textsc{Predicted} \textsc{Label}: contradiction & 
            \textsc{Predicted} \textsc{Label}: neutral       \\
            \textsc{Explanation}: A dog is not a bird. & 
            \textsc{Explanation}: A dog looking at water does not imply a bird is near water. \\ 
            \textsc{Extracted Knowledge}:  \{snow, RelatedTo, water\}, \{dog, DistinctFrom, cat\} & \textsc{Extracted Knowledge}: \{snow, RelatedTo, water\}, \{dog, DistinctFrom, cat\} \\
            \midrule

            \multicolumn{2}{c}{{\makecell{\textsc{Question}: What is a person who is good at sports considered? \\ \textsc{Choices}: talented, untalented, good at
            }}} \\ 

            \textsc{Choices}: talented, affluent, reproduce & \textsc{Choices}: talented, untalented, good at \\
            
            \textsc{Predicted Label}: talented & 
            \textsc{Predicted Label}: talented     \\
            \textsc{Explanation}: a person who is good at sports is considered talented. & 
            \textsc{Explanation}: a person who is good at sports is considered talented.  \\
            \textsc{Extracted Knowledge}: \{talent, RelatedTo, sports\} & \textsc{Extracted Knowledge}: \{talent, RelatedTo, sports\} \\
	\bottomrule
	\end{tabular}}}\vspace{-1ex}    
 \caption{Examples of successfully defended instances by the KnowNILE model on e-SNLI and by the KnowWT5 model on Cos-E. This table should be read together with Table \ref{table.attacked_samples_defended} to appreciate the defence. }\label{table.defended_examples_more}%
	\end{center}
\end{table*}

\begin{table*}[ht]
	\begin{center}
		\renewcommand{\arraystretch}{1.0}
		\footnotesize{
			\centering{\setlength\tabcolsep{5.0pt}
   
            \begin{tabular}{p{0.45\linewidth} | p{0.45\linewidth}}
  		\toprule
    
            \multicolumn{2}{c}{\makecell[c]{
            \textsc{Model}: KnowWT5 \\
            \textsc{Premise}: A young family enjoys feeling ocean waves lap at their feet.
            }} \\
             \makecell[c]{\textsc{Original}} & \makecell[c]{\textsc{Adversarial}} \\           
            \textsc{Hypothesis}: A family is at the beach.  & \textsc{Hypothesis}: The family is not at the beach.  \\
            \textsc{Predicted Label}: entailment & 
            \textsc{Predicted Label}: entailment \\
            \textsc{Explanation}: Ocean waves lap at their feet implies that they are at the beach. & 
            \textsc{Explanation}: Ocean waves lap at their feet implies that they are not at the beach. \\  
            & \textsc{Extracted Knowledge}: \{feet, HasA, fingerprints\} \\
            \midrule

            \multicolumn{2}{c}{\makecell[c]{
            \textsc{Model}: KnowNILE \\
            \textsc{Premise}: Some dogs are running on a deserted beach.
            }} \\
             \makecell[c]{\textsc{Original}} & \makecell[c]{\textsc{Adversarial}} \\           
            \textsc{Hypothesis}: There are multiple dogs present.  & \textsc{Hypothesis}: There are not many dogs on the beach.  \\
            \textsc{Predicted Label}: entailment & 
            \textsc{Predicted Label}: entailment \\
            \textsc{Explanation}: Some dogs implies multiple dogs. & 
            \textsc{Explanation}: Some dogs implies not many dogs. \\  
            & \textsc{Extracted Knowledge}: \{dog, DistinctFrom, cat\} \\
            \midrule

            \multicolumn{2}{c}{\makecell[c]{
            \textsc{Model}: KnowCAGE \\
            \textsc{Question}: What does the sky do before a rain?
            }} \\
            \textsc{Choices}: cloud over, rain water, appear blue & \textsc{Choices}: cloud over, appear blue, appear green \\
            \textsc{Predicted Label}: appear blue & 
            \textsc{Predicted Label}: appear green \\
            \textsc{Explanation}: the sky appears blue before a rain & 
            \textsc{Explanation}: the sky appears green before a rain \\
            & \textsc{Extracted Knowledge}: \{sky, UsedFor, rain\} \\

	\bottomrule
	\end{tabular}}}
     \caption{Examples of inconsistent NLEs detected by eKnowIA but not defended by Know-models. The extracted knowledge triplets are not highly related to generating correct explanations.}\label{table.not_defended_example}%
	\end{center}
	\vspace{-2ex}
\end{table*}

\begin{table*}[ht]
	\begin{center}
		\renewcommand{\arraystretch}{1.0}
		\footnotesize{
			\centering{\setlength\tabcolsep{5.0pt}
   
            \begin{tabular}{p{0.45\linewidth} | p{0.45\linewidth}}
  		\toprule
    
            \multicolumn{2}{c}{\makecell[c]{
            \textsc{Model}: KnowWT5 \\
            \textsc{Premise}: The collie is standing outdoors on a sandy area.
            }} \\
             \makecell[c]{\textsc{Original}} & \makecell[c]{\textsc{Adversarial}} \\           
            \textsc{Hypothesis}: The collie is standing in the sand.  & \textsc{Hypothesis}: The collie is standing on stone.  \\
            \textsc{Predicted Label}: entailment & 
            \textsc{Predicted Label}: entailment \\
            \textsc{Explanation}: A sandy area is made of sand. & 
            \textsc{Explanation}: A sandy area is made of stone. \\  
            & \textsc{Extracted Knowledge}: \{sand, RelatedTo, rock\} \\
            \midrule
    
            \multicolumn{2}{c}{\makecell[c]{
            \textsc{Model}: KnowNILE \\
            \textsc{Premise}: Coach talks with football player, other players and crowd in background.
            }} \\
             \makecell[c]{\textsc{Original}} & \makecell[c]{\textsc{Adversarial}} \\           
            \textsc{Hypothesis}: A football player is climbing into the stands at a game.  & \textsc{Hypothesis}: A football player talks to a crowd.  \\
            \textsc{Predicted Label}: contradiction & 
            \textsc{Predicted Label}: entailment \\
            \textsc{Explanation}: A coach is not a football player. & 
            \textsc{Explanation}: A coach is a football player. \\  
            & \textsc{Extracted Knowledge}: \{crowd, IsA, gathering\}, \{player, PartOf, team\}, \{football player, DerivedFrom, football\} \\

	\bottomrule
	\end{tabular}}}
     \caption{Examples of newly detected instances with inconsistent NLEs by eKnowIA for the \textsc{Know} models. The extracted knowledge triplets exhibit low relevance and confuse the model to generate incorrect explanations.}\label{table.newly_introduced}%
	\end{center}
	\vspace{-2ex}
\end{table*}

                     

\end{document}